\documentclass{article}

 \usepackage[final]{nips_2017}

\usepackage[utf8]{inputenc} 
\usepackage[T1]{fontenc}    
\usepackage{hyperref}       
\usepackage{url}            
\usepackage{booktabs}       
\usepackage{amsfonts}       
\usepackage{nicefrac}       
\usepackage{microtype}      
\usepackage{bbm}
\usepackage[cmex10]{amsmath}
\usepackage{amsthm}
\usepackage{wrapfig}
\usepackage{graphicx}
\newcommand{\supp}{\operatornamewithlimits{supp}}

\title{Kernels on fuzzy sets: an overview}

\author{
  Jorge.~Guevara\\
  IBM Research\\
  Sao Paulo, Brazil\\
  \texttt{jorgegd@br.ibm.com} \\
   \And
   Roberto Hirata Jr. \\
   University of Sao Paulo \\
   Sao Paulo, Brazil \\
   \texttt{hirata@ime.usp.br} \\
  \And
  Stéphane Canu \\
  INSA de Rouen \\
  Rouen, France \\
  \texttt{scanu@insa-rouen.fr}
}

\begin{document}

\maketitle

\begin{abstract}
  This paper introduces the concept of kernels on fuzzy sets as a similarity measure for $[0,1]$-valued  functions, a.k.a. \emph{membership functions of fuzzy sets}. 
  We defined the following  classes of kernels:  the cross product, the intersection, the non-singleton and the distance-based kernels on fuzzy sets.
  Applicability of those kernels are on machine learning and data science tasks where uncertainty in data has an ontic or epistemistic interpretation.
\end{abstract}

\section{Introduction}
%
Kernels on fuzzy sets were introduced by \citet{guevara2015modelos} as a mean to estimate a similarity measure between fuzzy sets with geometrical interpretation on Reproducing Kernel Hilbert Spaces.
Fuzzy sets are relaxed version of sets in the sense that fuzzy sets have $L$-valued  characteristic functions, where $L$ is a complete lattice, instead of  having $\{0,1\}$-valued functions. 
For instance, if we use the unit interval for $L$, it is possible to have a \emph{degree of membership} for elements in that kind of sets. In that sense,  a fuzzy set can be completely characterized by its membership function: $X:\Omega\to[0,1]$, and the evaluation $X(x)$ for some $x\in\Omega$ can be understood as the \emph{degree of membership} of $x$ to the fuzzy set with that membership function.
Fuzzy sets were introduced by 
Lotfi A. Zadeh in 1965
~\citet{DBLP:journals/iandc/Zadeh65} and that concept has been used since then in different areas of science.
%

The aim of this paper is introduce the concept of kernels on fuzzy sets to the machine learning community.
As all the computations are done using (membership) functions and some tools from fuzzy theory, we believe that this new tool would be helpful for machine learning and data science practitioners in problems  where data can be better modelled with that kind of structure.


\section{Why kernels on fuzzy sets?}
Fuzzy sets have been widely  used to model \emph{uncertainty}
in observational data, using either \emph{ontic} or \emph{epistemic} interpretation. Ontic, in the sense that point-wise uncertainties can be modeled by entities, i.e, FS can model set-valued attributes. Epistemistic, in the sense that a FS is a model for incomplete information on single-valued attributes, i.e., a model for non-precise data.
Using the ontic interpretation it is possible to think that fuzzy sets are elements with some underlying probabilistic law, and, hence it is possible to have concepts such as fuzzy-valued random variables \cite{kwakernaak1978fuzzy}.
Some modelling examples of ontic FS are: a region within an gray-scale image, a frequency profile, fuzzy clusters, a convolutional kernel on deep learning, etc.
From the epistemic point of view, FS can be used to model a region within images describing the no well-known location of an object, for example, a statement describing the (unknown) age of a person, a nested set of intervals containing some unknown deterministic value (\cite{hullermeier2005fuzzy,dubois2011ontic}). 
In practical applications, membership functions can be constructed very easy
using either arbitrary functions derived from  expert's knowledge for example,  
or using data-driven approaches (from histograms or quantile functions, for example) without assuming any probabilistic law for the data generation process.
However, there is a lack of use of fuzzy modeling techniques as an alternative tool from the ML community as it was noted by \cite{hullermeier2005fuzzy}. This research attempts to fill this gap in the kernel method area.
In this research we use the idea of kernels in order to estimate a similarity measure between fuzzy sets. This not only allow to have a geometric interpretation in Reproducing Kernel Hilbert Spaces for those similarity measures but also to use all the machine learning techniques from kernel methods on tasks where data can be modelled by fuzzy sets.

\section{Kernels}
Let $\mathcal{F}(\Omega)$ be the set of all FS with a membership function $X:\Omega\to[0,1]$
As  fuzzy sets are completely characterized by their membership functions, we will use the same capital letter for denoting either a fuzzy set or its membership function, i.e., $X(x)$ denotes the degree of membership of an element $x\in\Omega$ to a fuzzy set $X\in\mathcal{F}(\Omega)$. A \emph{kernel on fuzzy sets} is then  a real-valued mapping defined on $\mathcal{F}(\Omega)\times\mathcal{F}(\Omega)$.
In what follows we present four classes of kernels on fuzzy sets.

{\bf The cross  product kernel on fuzzy sets} - 
Let $k_1,k_2$ be  two real-valued kernels  defined on
$\Omega \times \Omega$ and
$[0,1] \times [0,1]$ respectively. The \emph{cross product} kernel on fuzzy sets is a function $k_{\times}:\mathcal{F}(\Omega)\times\mathcal{F}(\Omega)\to\mathbb{R}$ defined by:
\begin{equation}\label{eq:prodCruzEquiv}
k_{\times}(X,Y)=\sum_{\substack{x\in \supp(X),\\\ y\in \supp(Y)}}k_1\otimes k_2\big((x,X(x)),(y,Y(y))\big),
\end{equation}
where $X(x)$ and $Y(y)$ are the membership degrees for the elements $x,y\in\Omega$ to the fuzzy sets $X,Y$, the support of a fuzzy set is denoted by $\supp$, i.e. the set: $\{x\in\Omega \mid X(x)>0\}$ and
the \emph{tensorial product}: 
 $k_1\otimes k_2:(\Omega\times [0,1])\times (\Omega\times[0,1])\to\mathbb{R}$, is defined by:
 $k_1\otimes k_2\big(x,X(x),y,Y(y)\big)=k_1(x,y)\;  k_2(X(x),Y(y))$.
 Straightforward examples of positive definite cross product kernels on fuzzy sets can be obtained using positive definite kernels for $k_1$ and $k_2$, for example, if $k_2$ is always the linear kernel we have following kernels: $k_{\times}(X,Y)=\sum_{\substack{x\in \supp(X),\\\ y\in \supp(Y)}}xyX(x)Y(y)$, which uses a linear kernel for $k_1$. Also, if we set  $k_1$ to be the RBF kernel we have: $k_{\times}(X,Y)=\sum_{\substack{x\in \supp(X),\\\ y\in \supp(Y)}}\exp(-\gamma\|x-y\|^2)X(x)Y(y)$. Another example is given by defining the finite measure space $(\Omega, \mathcal{A},\mu)$ and
assuming that $k_1, k_2$  are continuous kernels functions with finite integral, then, 
the kernel $k_{\times}(X,Y)=\iint_{\substack{x\in \supp(X),\\\ y\in \supp(Y)}}k_1\otimes
  k_2\big((x,X(x)),(y,Y(y))\big)d \mu(x)d
  \mu(y)$, 
is a cross product kernel on fuzzy sets.
An instance of this kernel is given when we use a probability measure $\mathbb{P}$ instead of $\mu$, the resulting kernel incorporates two kinds of uncertainty modelling: fuzziness and randomness. 
Fuzziness in the form of membership functions and randomness
because, independently of the degree of membership of $x$ to the fuzzy
set $X$, the above formulation considers the values $x$ being outcomes
of a random variable with probability distribution $\mathbb{P}$.

The cross product kernel on fuzzy sets was presented by
\cite{8015459}. That kernel always would be positive definite if  $k_1$ and $k_2$ are positive definite. Kernel $k_{\times}$ is a natural extension of the kernel on sets to the fuzzy set domain. It can be shown that $k_{\times}$ is indeed a kind of convolution kernel (\cite{haussler1999convolution}), and that under some assumptions it embeds  probability distributions into RKHS. 
This kernel was successfully used in supervised classification on attribute noisy datasets, where it was shown that the kernel is resistant to injected random noise over the values of the predictors (see reference \cite{8015459} for the experiments).

{\bf The intersection kernel on fuzzy sets}, this kernel is based on the intersection operation between fuzzy sets. The main idea is to use the concept of \emph{finite decomposition} of sets within  a  semi-ring of sets $\mathcal{S}$, for our purposes we assume that the support of the fuzzy sets of interest is an element of a semi-ring of sets.
In order to define the intersection kernel on fuzzy sets we previously need the concept of semi-ring of sets:
a \emph{semi-ring of sets}, $\mathcal{S}\subseteq\Omega$, is a subset of  the power set  $ \mathcal{P}(\Omega)$,
satisfying the following conditions: 1) $\phi\in \mathcal{S}$, $\phi$ is the empty set,
2)  $A,B\in \mathcal{S} \implies A\cap B \in \mathcal{S}$, and 3)
 for all  $A, A_1 \in \mathcal{S}$ such that $A_1\subseteq A$,
  there is a sequence of pairwise disjoint sets: $A_2,A_3,\dots A_N\in\mathcal{S}$, such that: $A=\bigcup_{i=1}^N A_i$, this last
 condition 3 is known as the \emph{finite decomposition of } a set $A$.
\citet{gartner2008kernels}, shows that a kernel $k:\mathcal{S}\times\mathcal{S}\to\mathbb{R}$ defined by
$k(A,A')=\rho(A\cap A')$  is  positive definite, where 
$\rho:\mathcal{S}\to[0,\infty]$ is a measure defined on semi-ring of sets. We will use the same reasoning for defining a kernel based on the intersection of fuzzy sets. In that sense, we will denote by  $\mathcal{F}_{\mathcal{S}}(\Omega)$  the set of fuzzy sets whose support is an element of a semi-ring $\mathcal{S}$ and we will use the indicator function ${\bf 1}_{supp(X)}A$, that is one if $A\subseteq supp(X)$ and zero otherwise. 
Hence, a natural way to \emph{measure the support of a fuzzy set} is by using the measure $\rho$ (defined before) as follows: let denote by $\mathcal{A}\subseteq \mathcal{S}$ a finite system of pairwise disjoint sets and $\mathcal{B}\subseteq \mathcal{A}$, then the \emph{measure of the support of a fuzzy set} $X$ is defined by: $\rho(supp(X))=\sum_{A\in\mathcal{B}\subseteq\mathcal{A}}\rho(A)=\sum_{A\in\mathcal{A}}\rho(A){\bf 1}_{supp(X)}(A)$, where we used the fact that $supp(X)=\bigcup_{A\in\mathcal{B}\subseteq\mathcal{A}}A$.
All this analysis, allow us to have the following expression for \emph{measuring the support of the intersection of two fuzzy sets} $X,Y\in\mathcal{F}_{\mathcal{S}}(\Omega)$:
\begin{equation}\label{eq:medidaSuporte}
\rho(supp(X\cap Y))=\sum_{A\in\mathcal{A}}\rho(A){\bf 1}_{supp(X)}(A) {\bf 1}_{supp(Y)}(A).
\end{equation}
The intersection kernel on fuzzy sets is then the function:
$k_{\cap}:\mathcal{F}_{\mathcal{S}}(\Omega)\times\mathcal{F}_{\mathcal{S}}(\Omega)\to\mathbb{R}$,
satisfying:
\begin{equation}\label{eq:intersectionKernelFS}
            k_{\cap}(X,Y)= \sum_{A\in \mathcal{A}} \big(X\cap Y\big)(A)\rho(A) {\bf 1}_{supp(X)}(A) {\bf 1}_{supp(Y)}(A),
\end{equation}
where $\big(X\cap Y\big)(A)$ is an abuse of notation to indicate $\sum_{x\in A}\big(X\cap Y\big)\;(x)$, i.e., the total contribution of the membership degrees of elements belonging to $A$, evaluated in the membership function of $X\cap Y$.
Intersection of fuzzy sets are implemented via  \emph{T-norm} operators
which are mappings of the form $T:[0,1]^2\to[0,1]$
such that, for all $x,y,z\in[0,1]$, satisfy:
1) commutativity: $T(x,y)=T(y,x)$;
2) associativity: $T(x,T(y,z))=T(T(x,y),z)$;
3) monotonicity: $y\leq z \Rightarrow T(x,y)\leq T(x,z)$; and
4) limit condition $T(x,1)=x$.
(see ref. \cite{yu:generalized,klement2000triangular} for additionally definition and notations). Using a T-norm operator $T$, we have the following T-norm based kernel $k_{\cap}$:
$$k_{\cap}(X,Y)=\sum_{A\in \mathcal{C}_{X,Y}} \left (\sum_{x\in A}T(X(x),Y(x)) \right)\rho(A),$$
where, for ease of notation we use 
$\mathcal{C}_{X,Y}=\{A\in \mathcal{A}| {\bf 1}_{supp(X)}(A) {\bf 1}_{supp(Y)}(A) =1 \}$.
Table \ref{tab:kernel} shows several kernels  $k_{\cap}(X,Y)$ derived from common T-norms.
 \begin{table*}[hbt]
 \centering
 \begin{tabular}{lc}
\toprule
 Kernel $k_{\cap}$&T-norm\\
\midrule
  $k_{\cap\_\min}(X,Y)=\sum_{A\in \mathcal{C_{X,Y}}} \sum_{x\in A}\min(X(x),Y(x))\rho(A)$&minimum\\
\\
  $k_{\cap\_\text{pro}}(X,Y)=\sum_{A\in \mathcal{C_{X,Y}}} \sum_{x\in A}X(x)Y(x)\rho(A)$&product\\
 \\
  $k_{\cap\_\text{\L{}uk}}(X,Y)=\sum_{A\in \mathcal{C_{X,Y}}} \sum_{x\in A}\max(X(x)+Y(x)-1,0)\rho(A)$&\L{}ukasiewicz\\
\\
  $k_{\cap\_\text{Dra}}(X,Y)=\sum_{A\in \mathcal{C_{X,Y}}} \sum_{x\in A}Z(X(x),Y(x))\rho(A)$&Drastic\\
\midrule
\end{tabular}
\caption{Different formulations for $k_{\cap}$ induced by different T-norms operators}
\label{tab:kernel}
\end{table*}

This kernel was presented in \cite{6891628}, this kernel is   positive definite if the T-norm $T$ is a positive definite function.

{\bf The non-singleton kernel on fuzzy sets}, 
this kernel is  a function
$\mathcal{F}(\Omega)\times\mathcal{F}(\Omega)\to[0,1]$ defined by:
\begin{equation}
k_{nsk}(X,Y)=\underset{x\in \Omega}{\sup} \left (T(X(x),Y(x)) \right),
\end{equation}
where $T$ is an T-norm operator, and $sup$ is the \emph{supremum}.
That kernel is also a kernel based on the intersection of fuzzy sets, because T-norms are used to estimate the intersection between fuzzy sets. In this sense a more general definition for this kernel is given by: $k_{nsk}(X,Y)=\underset{x\in \Omega}{\sup} \;\;\; \big(X\cap Y\big) (x)$. This kernel was derived from the analysis of the interaction between non-singleton fuzzy systems and its inputs in the context of fuzzy inference, see \cite{6622409} for details of that analysis. 
Particularly, 
for two tuples of fuzzy sets: $X=(X_1,\dots,X_d,\dots,X_D)$ and $Y=(Y_1,\dots,X_d,\dots,Y_D)$,
with  Gaussian membership functions, i.e. $[0.1]$-valued functions taken the following form: 
$X_d(.)=\exp\left(-\frac{1}{2}\frac{(.-m_d)^2}{\sigma_d^2}\right)$,
where,
 $m_d\in\mathbb{R}$ amd $\sigma_d\in\mathbb{R}^+$ are the function parameters.
Then, we proved that 
\begin{equation}\label{eq:kernelGaussianoTSKParametro}
k_{nsk}^{\gamma}(X,Y)= \prod_{d=1}^D \exp\left( -\frac{1}{2}\frac{(m_d-m'_d)^2}{\sigma_d^2+(\sigma'_d)^2}\right),
\end{equation}
is a positive definite kernel.
More instances of this kernel can be found in \cite{6622409}.
Another important results regarding those kernels are that 
such kernels are fuzzy equivalence relation w.r.t a T-norm operator (Corollary $6$ in \cite{Moser20061787}), they are
at least $T_{cos}$-transitive (\cite{Moser20061787}) and they can be interpreted as
fuzzy logic formulas for fuzzy rules (Theorem $9$ in 
\cite{Moser:2006:RGK:1248547.1248640}).
This kernel was applied on supervised classification of data containing interval-valued predictors.

{\bf Distance-based kernels on fuzzy sets}, this kernels are based on the concept of \emph{distance substitution kernels} (\cite{haasdonk2004learning}). The main ideia is to use metrics, pseudo-metrics or semi-metrics in order to define symmetric kernels. For a metric $D$, and for $x,y\in\Omega$ \citet{haasdonk2004learning} defined  $\langle x,y \rangle_D^{x_0}=\frac{1}{2} \big( D(x,x_0)^2 +D(y,x_0)^2 - D(x,y)^2 \big )$, where $x_0$ is some arbitrary point in $\Omega$. We use the same idea to define the operation $\langle X,Y \rangle_D^{X_0}$ in a similar way for $X,Y\in\mathcal{F}(\Omega)$. Then, the following kernels are positive definite if $D$ is a metric between fuzzy sets:
1) $K(X,Y)=\langle X,Y \rangle_D^{X_0}$, which can be viewed as a kind of inner product kernel, 2) $K(X,Y)=\big (\alpha +\gamma \langle X,Y \rangle_D^{X_0}  \big )^\beta$, where $\alpha,\gamma\in\mathbb{R}^+$, $\beta\in\mathbb{N}$, and can be viewed as a polynomial type kernel,  and 3) $K(X,Y)=\exp(-\gamma D(X,Y)^2)$ which is a kind of Gaussian kernel. For instance, using the following metric on fuzzy sets: $D(X,X')=\dfrac{\sum_{x\in\Omega} | X(x)-X'(x)|}{\sum_{x\in\Omega} | X(x)+X'(x)|}$ and by inserting that metric into the kernel definition, i.e. $K_{D}(X,X')=\exp(-\lambda D(X,X')^2)$, we will have a positive definite kernel.
Further, if $D$ is not a metric but instead is a semi-metric or pseudo-metric, still it is possible to perform machine learning on symmetric kernels
(\cite{1030883,chapelle1999support, haasdonk2002tangent, moreno2003kullback}).
Some popular distances between fuzzy sets that could induce new kernels on fuzzy sets can be found in  \cite{bloch1999fuzzy,rosenfeld1985distances,chaudhur1996metric,diamond1994metric}.
Distance-based kernels on fuzzy sets were applied on two-sampled hyphotesis testing on heterogeneous data (\cite{FuzzSim}).



\section{Conclusions}
In this paper we introduced the concept of kernels on fuzzy sets, we presented four classes of that kind of kernels:  the cross product kernel on fuzzy sets that is an extension of the widely-known kernel on sets to the fuzzy set domain; the intersection kernel on fuzzy sets that uses some concepts from set and fuzzy set theory for its own definition; the non-singleton kernel on fuzzy sets that was basically derived from the analysis of non-singleton fuzzy systems; and the distance-based kernels on fuzzy sets that uses the concept of distance substitution kernels.
We think that that class of kernels are usefully in contexts where data uncertainty has an ontic or epistemic interpretation. There are some successfully applications of those kernels in tasks like classification of attribute noisy data, classification of interval data and kernel hypothesis testing. However, we think that more experimental research must be done using those kernels  in order to validate or extrapolate their applicability.

\subsubsection*{Acknowledgments}
The authors are thankful with
FAPESP grant \# 2015/01587-0, CNPq, CAPES, NAP eScience - PRP - USP
and IBM Research Brazil
for their financial support.

\medskip

\small
\bibliographystyle{plainnat}
\bibliography{main}

\end{document}